\documentclass[sigconf]{acmart}
\usepackage{bm}
\usepackage{xcolor}
\usepackage{placeins}
\usepackage{multirow}
\usepackage{todonotes}

\AtBeginDocument{%
  \providecommand\BibTeX{{%
    \normalfont B\kern-0.5em{\scshape i\kern-0.25em b}\kern-0.8em\TeX}}}
    
\setcopyright{none}
\renewcommand\footnotetextcopyrightpermission[1]{} 
\begin{document}
\sloppy

\title{Robust Document Representations using Latent Topics and Metadata}

\author{Natraj Raman}
\email{natraj.raman@jpmorgan.com}
\affiliation{%
  \institution{J.P. Morgan AI Research}
  \city{London}
 \country{U.K}
}

\author{Armineh Nourbakhsh}
\email{armineh.nourbakhsh@jpmorgan.com}
\affiliation{%
  \institution{J.P. Morgan AI Research}
  \city{New York}
 \country{U.S.A}
}

\author{Sameena Shah}
\email{sameena.shah@jpmorgan.com}
\affiliation{%
  \institution{J.P. Morgan AI Research}
  \city{New York}
 \country{U.S.A}
}

\author{Manuela Veloso}
\email{manuela.veloso@jpmorgan.com}
\affiliation{%
  \institution{J.P. Morgan AI Research}
  \city{New York}
 \country{U.S.A}
}

\renewcommand{\shortauthors}{Raman, et al.}

\begin{abstract}
Task specific fine-tuning of a pre-trained neural language model using a custom softmax output layer is the de facto approach of late when dealing with document classification problems. This technique is not adequate when labeled examples are not available at training time and when the metadata artifacts in a document must be exploited. We address these challenges by generating document representations that capture both text and metadata artifacts in a task agnostic manner. Instead of traditional auto-regressive or auto-encoding based training, our novel self-supervised approach learns a soft-partition of the input space when generating text embeddings. Specifically, we employ a pre-learned topic model distribution as surrogate labels and construct a loss function based on KL divergence. Our solution also incorporates metadata explicitly rather than just augmenting them with text. The generated document embeddings exhibit compositional characteristics and are directly used by downstream classification tasks to create decision boundaries from a small number of labeled examples, thereby eschewing complicated recognition methods. 
We demonstrate through extensive evaluation that our proposed cross-model fusion solution outperforms several competitive baselines on multiple datasets. 

\end{abstract}

%
%
%

\maketitle
\pagestyle{plain} 

\section{Introduction}
The current popularity of transformer-based models in Natural Language Processing is owed to their capacity in constructing semantically rich representations and to their ability to accommodate transfer-learning. This enables users to pre-train transformer-based language models on large unlabeled corpora, and then fine-tune the representations using smaller sets of labeled examples \cite{howard2018}. How small the labeled dataset can be, depends on the complexity of the task, the domain-similarity between the pre-training and labeled datasets, and the model architecture\cite{aharoni2020}. Recent research in few-shot learning has focused largely on the last aspect, proposing neural architectures that capture semantic and compositional information efficiently and allow for faster convergence on new tasks \cite{zhang2020few}. However, the state-of-the-art research is still far from human performance on many tasks \cite{schick2020}. This is even more pronounced in zero-shot experiments, where no labeled data is available at training time \cite{brown2020gpt3}. 

Due to the emphasis on scalability and generalizability of these models, what is often left out of consideration is the practical aspects of how these models are commonly applied in real-world settings. As an example, datasets are very often accompanied by metadata tags that include some signal about the nature and content of each document in the corpus. Corporate communications, financial reports, regulatory disclosures, policy guidelines, Wikipedia entries, social media messages, and many other forms of textual records often bear tags indicating their source, type, purpose, or an enterprise categorization standard. 
This metadata is often stripped before models are applied to the text, in order to avoid convoluted and bespoke architectures. In some cases, the metadata is simply concatenated to the document \cite{zhang2020minsup} without specific controls on how representations are generated from raw text versus metadata tags. A flexible solution that can generalize to various types of metadata can prove useful in augmenting the richness of semantic representations. 

\begin{figure*}[htbp]
\centering
\includegraphics[width=\textwidth]{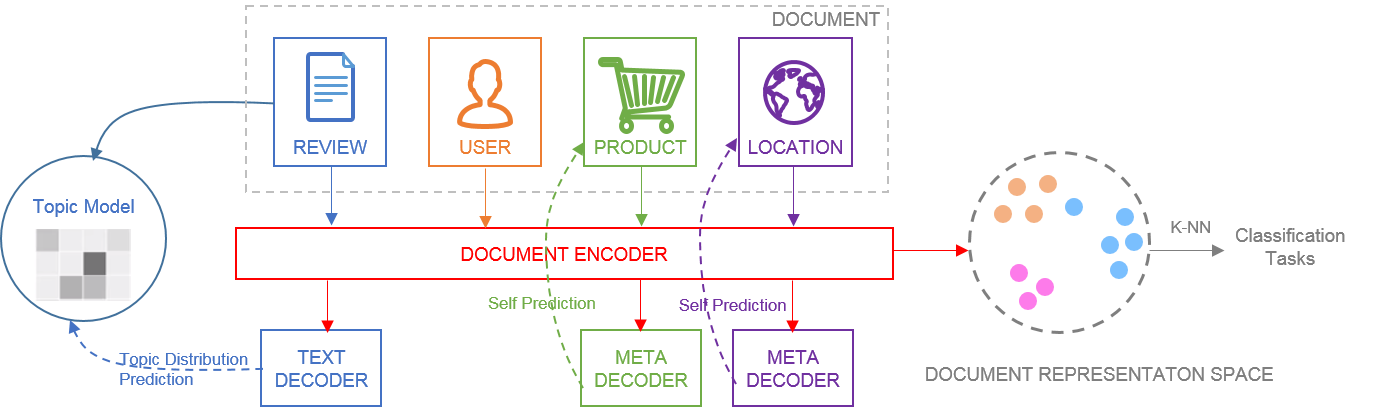}
\caption{Approach Overview. Multi-part input document is converted into an embedding representation using a neural encoder. Self-supervision during training is based on latent topic distribution and (optional) reconstruction for metadata. Input points are then classified based on its neighborhood in the representation space.} 
\label{fig_overview}
\end{figure*}

Another aspect that current research often leaves out is the topic distribution of the unlabeled dataset. Transformer models are commonly trained in a stochastic fashion, where the global composition of topics in the corpus is not explicitly built into the loss function. Often the pre-training corpus is large enough for this effect to likely be negligible, but when the corpus is smaller than the common multi-million-document setting, this lack of insight into global distributional statistics may have an impact on the compostionality of the resulting representations.

In this study, we explore how addressing both of these issues can improve the performance of transformer-based models on document classification tasks. We propose a simple yet effective framework that encapsulates universal distributional statistics about the raw text, as well as encodings for metadata tags. The framework includes several components that bring together useful characteristics of self-supervised and multi-task learning:
\begin{enumerate}
    \item An LDA-based topic model \cite{blei2003latent} is used to train a transformer-based language model to effectively encode distributional characteristics of the unlabeled dataset. This is done by modeling the text in each document as a distribution of topics, and training the transformer on a KL-Divergence loss against those distributions.
    \item Metadata artifacts are used to enrich document representations. Each metadata tag is encoded separately and concatenated with the text representation. The encoding objective is adjusted based on the type of metadata.
    \item A multi-task objective is used to jointly learn text and metadata representations. 
    \item The resulting document representations show strong compositionality and can be plugged into a K-NN algorithm for document-level classification tasks. 
\end{enumerate}

Figure \ref{fig_overview} illustrates our framework when applied to a hypothetical dataset of product reviews with a diverse set of metadata artifacts. The raw text of each review is paired with other metadata artifacts available, such as user profile information, product identifiers, and location information. All the artifacts are fed into a deep learning model with the self-learning mechanism adjusted to match the type of artifact presented. For example, for the raw text of the reviews, instead of using a standard Masked Language Model (MLM) objective \cite{devlin2018bert}, the model learns by predicting the latent topic distribution of the text based on a pre-trained generative model such as LDA \cite{blei2003latent}. For categorical metadata such as product identifiers and location, the model predicts the specific metadata tag. Certain metadata can also be directly encoded, bypassing the self-learning task. The resulting representations are semantically rich, and can be plugged into a simple K-NN model for various label-prediction tasks, bypassing the need for complicated, task-specific classification models. 

We demonstrate how the representations created by our framework exhibit compositional characteristics that can be useful to granular classification tasks. While small and scalable to many different settings, our framework improves the performance of transformer-based language models on classification tasks on a variety of datasets. Our experiments show that regardless of the underlying neural architecture, performance is enhanced by a robust minimum of 5\% over a conventional fine-tuned model that uses the special CLS embedding for document classification. We also explore the robustness of the framework through a series of ablation studies. The remaining sections of this paper lay out our methodology, describe our datasets, and present experimental results.

\section{Related Work}\label{sec_relwork}
Semi-supervised learning in neural language models has largely focused on ``pre-train and fine-tune'' pipelines \cite{howard2018}, which take advantage of large unlabeled datasets, paired with small labeled datasets. Having produced rich representations during the pre-training phase, the model leverages the labeled dataset to calibrate its parameters in an inductive fashion against a new task \cite{devlin2018bert}. To keep the models simple and flexible, the fine-tuning process needs to be parameter-efficient, fast, and amenable to ``plug-and-play'' applications. This can prove difficult when the distributions of labeled and unlabeled datasets diverge \cite{aharoni2020} or when the labeled dataset is prohibitively small \cite{brown2020gpt3}. A large body of research has thus focused on few-sample learning, attempting to make the pre-training process more robust against such issues \cite{sharaf2020metalearning}. However, performance on NLP tasks remains far from human baselines \cite{wang2020few}. This is partly due to the fact that the inductive fine-tuning process fails to directly take advantage of universal distributional characteristics of the dataset beyond what is encoded in the pre-trained representations. Since the pre-trained representations are themselves stochastically generated, they are not guaranteed to seamlessly encode universal statistics. As a result in many applied studies the representations are paired with other distributional signal such as topic models and tf-idf vectors \cite{lim2020uob}. 

In response, a growing body of literature in recent years has focused on transductive learning approaches, especially in the computer vision domain \cite{lu2020few}. This paradigm allows the model to actively take advantage of sample distributions during inference. In natural language processing, this paradigm has been used to improve performance on tasks such as cross-domain text classification \cite{ionescu2018transductive} and neural machine translation \cite{poncelas2019transductive}. Transductive Support Vector Machines \cite{joachims1999transductive} have also been applied to granular classification tasks \cite{selvaraj2012extension}. These studies take advantage of a robust labeled dataset to scale to unseen datasets or domains. However, they do not address cases where the original dataset lacks enough labeled examples. Similarly, multi-task learning studies have addressed cases where the model can be robustly trained for one task such as entity extraction, and scale to other tasks such as co-reference resolution \cite{sanh2018hierarchical}.

In this study, we propose a transductive framework that can take advantage of a limited labeled dataset paired with a larger unlabeled dataset to generate rich representations for document classification tasks. The framework brings together useful characteristics of the above-mentioned approaches in a unique pipeline that leverages latent topic models and encodings of useful metadata via a multi-task loss.

\section{Model}\label{sec_model}

In traditional text classification tasks, the input consists of labeled pairs of training examples. The text is fine-tuned on a pre-trained model such as \cite{devlin2018bert} using a classification specific loss function based on these training labels. Our problem is different from the traditional setting on three aspects - (a) there are no labels available during training, (b) the model must be shared across different types of classification tasks and (c) the input contains metadata in addition to plain text. 

\begin{figure*}[htbp]
\centering
\includegraphics[height=7cm]{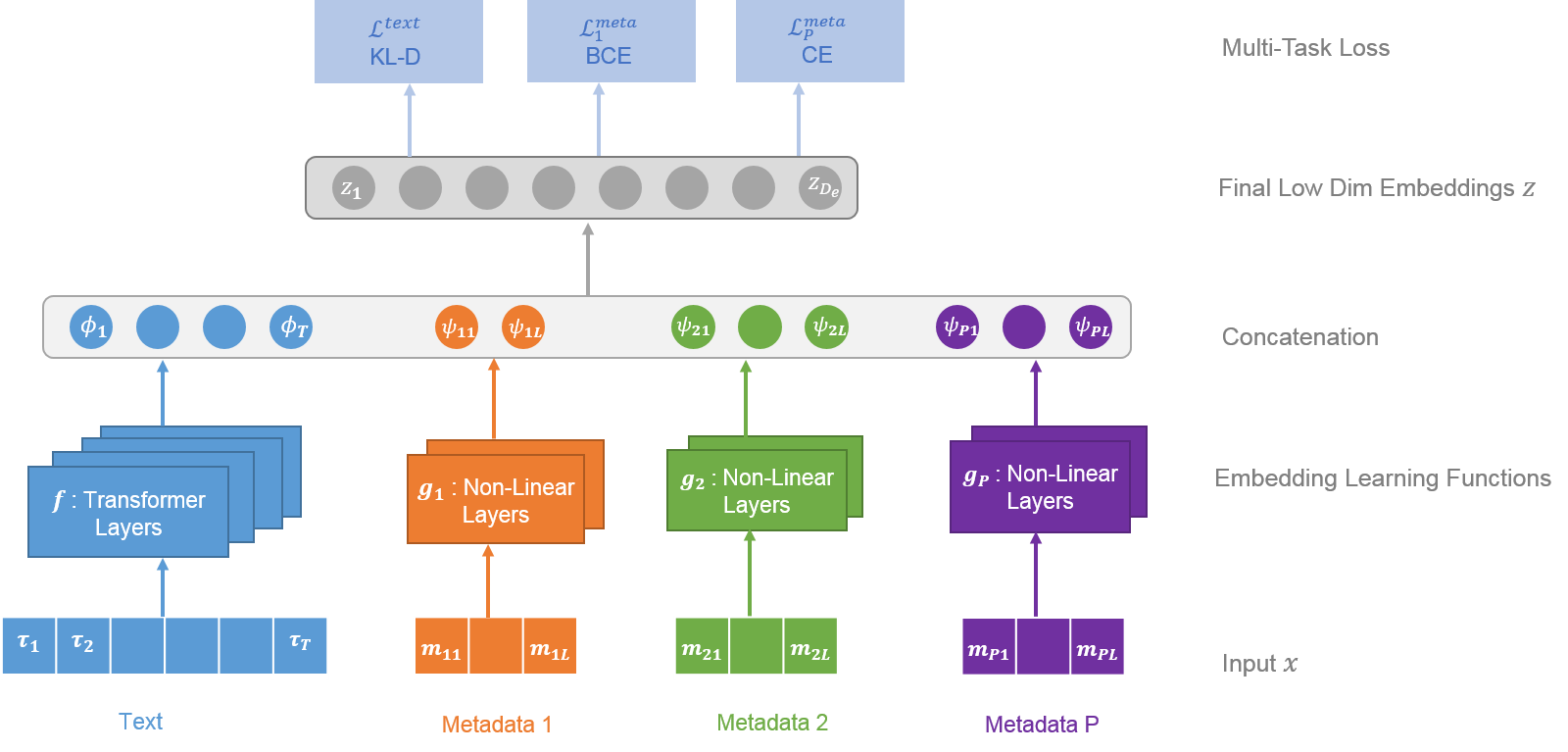}
\caption{Model Architecture. Embeddings learned independently for different input types are combined and then projected to a lower dimensional space. The model is trained using a multi-task objective function.} 
\label{fig_model1}
\end{figure*}

We first introduce the encoder architecture that is used to obtain an input representation that captures both the text and metadata information in a task agnostic manner. Then we present the decoder structure that employs self-supervision to define the training objective.  Finally, we  discuss how the learned representation is directly used in downstream classification tasks. Figure ~\ref{fig_model1} provides an overview of our model.

\subsection{Representation Learning}\label{subsec_rep}

Given $N$ training examples $X=\{x^1,...,x^N\}$, let $x=(\tau,m_1,...,m_P)$ be an input that contains text $\tau$ accompanied with ${P}$ different metadata artifacts ${m}$. The text consist of $T$ tokens $(\tau_1,...,\tau_T)$ from a fixed vocabulary and each metadata $m_p$ is a sequence $(m_{p1},...,m_{pl},...m_{pL})$ of fixed length $L$. Sequences shorter than $T$ or $L$ are simply padded. Let  $m_{pl} \in \Omega^p$, where $\Omega^p$ is the discrete set of information for $p^{th}$ metadata. 

The text input is converted into an interim embedding representation $\phi$ using a function
\begin{align}\label{eq_1}
    \textbf{f} : \tau \rightarrow \phi,\qquad \phi \in \mathbb{R}^{D_t}
\end{align}
where $D_t$ is the text embedding size. The function $\textbf{f}$ is a Transformer \cite{vaswani2017attention} model that employs a large number of layers with self-attention mechanism to capture dependencies between arbitrary positions of text in an efficient manner. The model is initialized with pre-trained parameters, thereby incorporating prior knowledge gained from training on large text corpora. The input tokens are augmented with a special token $[CLS]$ that represents the aggregate information of the entire sequence. The output corresponding to this token at the last layer is used as $\phi$. 

There are $P$ independent non-linear functions to convert each metadata input into an interim embedding representation $\psi$:
\begin{equation}\label{eq_2}
    \textbf{g}_p : m_{pl} \rightarrow \psi_{pl},\qquad \psi_{pl} \in \mathbb{R}^{D_p},\quad \forall p=1...P,l=1...L
\end{equation}
where $D_p$ is the metadata embedding size. We make use of a feed-forward network with multiple layers as the conversion function $\textbf{g}$, with each layer comprising of a linear transformation followed by a non-linear activation. If a metadata cannot be meaningfully interpreted (e.g. product code or user id), we use one-hot encoding of the metadata values as input. Otherwise, the input is set to a 300 dimensional vector derived by averaging the Glove \cite{pennington-etal-2014-glove} vectors corresponding to the words in metadata text. The embedding for a metadata sequence is aggregated using a mean function that masks out padded positions $\gamma \in \{0,1\}$ as
\begin{equation}
    \psi_p = \frac{1}{L}\sum_l\gamma_{pl}{\psi_{pl}}.
\end{equation}

The final embedding representation for an input $x$ is obtained by first concatenating the text and metadata embeddings and then projecting them to a lower-dimensional space using a linear transformation as follows:  
\begin{equation}\label{eq_4}
    z = W_z^\intercal(\phi \oplus \psi_1 \oplus ... \oplus \psi_P),\qquad z \in \mathbb{R}^{D_e}
\end{equation}
where $z$ is the final input embedding of size $D_e$ and $W_z \in \ \mathbb{R}^{(D_t+\sum_p D_p)\times D_e}$ is a parameter matrix.

\subsection {Self-supervised Loss}
In the absence of external supervisory information, self-supervision has emerged as a promising solution. The key idea here is to  generate synthetic labels automatically from the data and use these labels to construct loss functions. For text input, typically, the identities of some words in the text are masked and the model is trained to recover the original input. 

As an alternative, we propose a cross-model fusion approach. A topic model for the text corpora is learnt in an unsupervised manner and the inferred topic distribution is used as synthetic labels. By discovering latent semantics embedded in the text, topic models introduce a partition of the input space. Importantly, the mixed membership of topics for an input enables a soft partition rather than a hard partition. Such a fuzzy clustering approach is preferable since it can handle overlapping boundaries and complex structures in a better manner. This partition based training objective promotes separation of input data points into groups of similar points. Thus these points can now be classified by simply examining their neighborhood.

Formally, the topic distribution $\varphi$ corresponding to an input text is obtained using a function
\begin{align}
    \textbf{h} : \tau \rightarrow \varphi,\qquad \varphi \in \mathbb{R}^{K}
\end{align}
where $K$ is the number of topics and $\textbf{h}$ is a topic modeling function based on LDA \cite{blei2003latent}. The input embeddings obtained using \eqref{eq_4} are first linearly projected into the topic space as:
\begin{align}\label{eq_6}
    \lambda^n = W_t^\intercal z^n,\qquad \lambda^n \in \mathbb{R}^K.
\end{align}
Here $\lambda^n$ is the projected distribution in topic space for the $n^{th}$ training example and $W_t \in \mathbb{R}^{D_e \times K}$ is a parameter matrix. The Kullback-Leibler (KL) divergence between the pre-learned topic distribution and the input projection is minimized during training. This translates into the following loss function for text inputs:  
\begin{align}\label{eq_L1}
    \mathcal{L}^{text} = \sum_n \sum_k \varphi^n_k\:log\: \frac{\varphi^n_k}{\lambda^n_k}.
\end{align}

For the metadata, we employ a loss function that aims to minimize the reconstruction error between the input metadata value and the value decoded from the final embedding representation. Given that each metadata is a sequence of values from a discrete set, a multi-label binary cross-entropy loss is an appropriate choice. 

Let there be $V^p$ possible values for the $p^{th}$ metadata i.e. $|\Omega^p|=V^p$ and let $y_p \in \{0,1\}^{V^p}$ denote the multi-label values consolidated from an input metadata sequence\footnote{For instance, if the dataset is a collection of news articles, then the $p^{th}$ metadata artifact might indicate the set of countries related to a given news article. In that case $\Omega^p$ would be the set of all possible countries, $V^p$ would be the number of possible countries, and each $y_p$ would be a vector of size $V^p$ where indices corresponding to relevant countries are set to 1.}. A linear transformation decoder layer first converts the input embeddings into the metadata space as
\begin{align}
    \zeta^n_p = W_p^\intercal z^n,\qquad \zeta^n_p \in \mathbb{R}^{V^p}
\end{align}
where  $\zeta^n_p$ is the projection for the $n^{th}$ input and $W_p \in \mathbb{R}^{D_e \times V^p}$ is a parameter matrix as before. The reconstruction loss function is formulated as:
\begin{align}\label{eq_L2}
    \mathcal{L}^{meta}_{p} = \sum_n \sum_v -y_{p,v}\:log\:\sigma(\zeta^n_{p,v})-(1-y_{p,v})\:log\:(1-\sigma(\zeta^n_{p,v}))
\end{align}
where $\sigma$ is the standard sigmoid function.

Using the text and metadata losses in  ~\eqref{eq_L1} and ~\eqref{eq_L2}, the training objective is framed as   
\begin{align}\label{eq_L10}
    \min_{\substack{\theta}}\;\omega^{text} \mathcal{L}^{text} + \sum_p \omega^{meta}_{p}\mathcal{L}^{meta}_{p}
\end{align}
where $\theta$ is the set of all model parameters and $\omega$ is a real-valued hyper-parameter that controls the relative importance between the text and various metadata.

\subsection {Unseen Classification Tasks}
The input representations obtained using the above model have several desirable characteristics: 
\begin{itemize}
\item the embeddings are in a compact form because of the projection into a lower-dimensional space,
\item the salient information in the inputs are preserved by the reconstruction loss, and 
\item similar points are grouped together with the use of soft-partition labels.
\end{itemize}

Hence these embeddings can be used as-is in a non-parametric setting for downstream classification tasks. This approach is particularly attractive in situations where it is expensive to train new classification models or it may not be possible to perform training due to the scarcity of labeled examples.

Given a small number of labeled examples, we use nearest neighbor technique to compute the classification label of a query point. Specifically, the embeddings of the query point and the labeled points are obtained as outlined in Sec. \ref{subsec_rep} and the Euclidean distance between them is used to identify the nearest neighbors. The query point's class label is computed using a mode function on the nearest neighbor labels.

\section{Experiments}\label{sec_eval}
Experiments are conducted on three real-world datasets that vary in text style and nature of side information. We demonstrate below using these datasets that the proposed solution outperforms standard self-supervised training objectives for a variety of benchmark transformer models.    

\begin{table*}[htbp]
\caption{Data samples}
\label{tab_dataset}
\centering
\begin{tabular}{|l|l|r|}
\hline
\multirow{4}{*}{Github-AI} & \multicolumn{1}{l|}{Description} & \multicolumn{1}{l|}{ Handbag GAN. In this project I will implement a DCGAN to see if I can generate handbag designs...} \\\cline{2-3}
                                 & \multicolumn{1}{l|}{Repo Name} & \multicolumn{1}{l|}{GAN experiments} \\\cline{2-3}
                                 & \multicolumn{1}{l|}{Tags} & \multicolumn{1}{l|}{gan,dcgan,deep-learning,google-cloud} \\\cline{2-3}
                                 & \multicolumn{1}{l|}{Labels} & \multicolumn{1}{l|}{Image Generation (granular), Computer Vision (coarse)} \\\hline

\multirow{3}{*}{Amazon} & \multicolumn{1}{l|}{Review} & \multicolumn{1}{l|}{Best little ice cream maker works well, not too noisy, easy to clean. Recommend buying extra...} \\\cline{2-3}
                                 & \multicolumn{1}{l|}{Product} & \multicolumn{1}{l|}{B00000JGRT} \\\cline{2-3}
                                 & \multicolumn{1}{l|}{Label} & \multicolumn{1}{l|}{Home\_and\_Kitchen} \\\hline
\multirow{3}{*}{Twitter} & \multicolumn{1}{l|}{Tweet} & \multicolumn{1}{l|}{greek yogurt fresh fruit honey granola healthy living bakery.} \\\cline{2-3}
                                 & \multicolumn{1}{l|}{Hashtags} & \multicolumn{1}{l|}{\#healthyliving, \#bakeri} \\\cline{2-3}
                                 & \multicolumn{1}{l|}{Label} & \multicolumn{1}{l|}{Food} \\\hline
\end{tabular}
\end{table*}

\subsection{Datasets}\label{subsec_dset1}
The datasets correspond to three different domains and are publicly available. Table \ref{tab_dataset} lists a few sample data.

\textbf{Github-AI \cite{zhang2019higitclass}:}  This dataset contains a list of source code repositories that implement algorithms for various machine learning tasks. We use the description summary of a repository as text data. The repository name and the tagged keywords constitute metadata. Note that there can be multiple tags for a repository, and we use up to 5 tags as part of the metadata sequence. If there are fewer than 5 tags (or if the tags are absent), then the sequence is padded. The dataset also contains two different label fields for each repository that relate to the machine learning task hierarchy. The first set of labels denote the task domain in a coarse manner (e.g. \emph{Computer Vision}) while the second set of labels define a granular domain (e.g. \emph{Object Detection, Semantic Segmentation}, etc.). These labels are employed to evaluate the learned input representation for classification.   

\textbf{Amazon \cite{mcauley2013hidden}:}  Review data from online retailer Amazon is collected in this dataset. There are over 35 million reviews and we sample 10,000 reviews similar to the procedure in \cite{zhang2020minimally}. The comments entered by a reviewer is used as the text data while the product identifier is used as side information. Since there is a single unique product identifier for each review, the metadata sequence degenerates to a scalar in this case. The product category (e.g. \emph{books}, \emph{sports}, etc.) is used as the classification label.

\textbf{Twitter \cite{zhang2017react}:}  This social media dataset contains tweets collected during a three month period in 2014. The hashtags corresponding to a tweet (e.g. \emph{\#delicious, \#gym,} etc.) are used as side information. These tags do not conform to normal English style text with complex phrases and unstructured expressions. Hence we tokenize and segment \cite{baziotis2017} these tags so that a tag such as \emph{\#currentsituation} or \emph{\#makemyday}  is meaningfully interpreted as \emph{current situation} or \emph{make my day}. The tweet itself is used as the text data. Each tweet is associated with a category (e.g. \emph{food}, \emph{nightlife} etc.) which we use as its classification label.

\subsection{Baselines for Comparison}\label{subsec_baseline}
For comparison purposes, we use five state-of-the-art neural language models namely BERT~\cite{devlin2018bert}, DistilBERT~\cite{sanh2019distilbert}, XLNet~\cite{yang2019xlnet}, RoBERTa~\cite{liu2019roberta} and Electra~\cite{clark2020electra}. All these models are based on the Transformer~\cite{vaswani2017attention} architecture - however, they vary in training procedure. BERT masks the identities of some input tokens and aims to reconstruct the original tokens, while XLNet uses a generalized auto-regressive setup. Electra on the other hand replaces certain input tokens with plausible alternatives sampled from a generator network. RoBERTa and DistilBERT optimize the BERT training procedure with improved choice of hyper-parameters and parameter size reduction respectively.

There are three different evaluation setups corresponding to these models:   
\begin{itemize}
\item \textbf{No Finetuning:} The side information such as repository name or hashtag is augmented with text data to create a single text block. This text block is then used as input to the model for inference, and the $[CLS]$ token's embeddings from the last layer is used as the input representation.
\item \textbf{LM Finetuning:} As in the previous \emph{No Finetuning} setup, a single text block is created from the multi-part input. However, the language model is finetuned using this text before performing inference. This allows the model to adapt its weights based on the nature of the data. Finally, the input representation is inferred from the $[CLS]$ token of this finetuned model.
\item \textbf{Our Approach:} This setup reflects the architecture described in Section \ref{sec_model}. Text and side information are treated independently, with each input part being encoded by its corresponding embedding learner. Unlike the auto-encoding or auto-regressive training objective used in the \emph {LM Finetuning} setup, the transformer weights are updated by the custom self-supervised loss in equation  ~\eqref{eq_L10}.
\end{itemize}

\subsection{Settings}\label{subsec_settings}
We employ the base configuration of a transformer model, which typically has $12$ layers and $768$ hidden neurons per token. This allows the models to be trained using a single GPU instance (Tesla V100-SXM2-16GB).
The models are implemented using the pytorch version of Transformers library \cite{Wolf2019HuggingFacesTS}.  We use Adam Optimizer with an initial learning rate of $5e-5$ and an epsilon of $1e-8$. Other parameters include a drop out probability of $0.1$, sequence length of $512$ and batch size of $8$. The models were trained for $3$ epochs on the GitHub-AI dataset and for a single epoch on the Amazon and Twitter datasets to avoid overfitting. All hyper-parameters were chosen after a careful grid search.

For the metadata embedding learner, $tanh$ is used as the activation function. One-hot encoded input vectors are used for the tags in GitHub-AI dataset and product identifier in Amazon dataset while the repository name in GitHub-AI dataset and hashtags in Twitter dataset are initialized with Glove vectors as outlined in Section \ref{subsec_rep}.  The metadata embedding size $D_p$ is set to $50$ while the final input embedding size $D_e$ is set to $500$. 

When using nearest neighbor classification, only 10\% of the data is used as exemplars and the rest of the data is used for evaluation. The number of neighbors is set to $10$, and all points in the neighborhood are weighted equally. 

\subsection{Results}\label{subsec_results}
The classification results for GitHub-AI dataset is shown in Table \ref{tab_ghubaisub} and \ref{tab_ghubaisuper}. The former contains F1 scores for $14$ granular class labels while the latter is for $3$ coarse class labels. The granular classification task is particularly challenging due to the large number of classes and a small sample size. However, this reflects a typical real-world few-shot learning scenario. We see that there is no big difference in performance between \emph{No Finetuning} and \emph{LM Finetuning} setups. This dataset is very small with only $1600$ examples and hence the pre-trained weights dominate even after fine-tuning. In contrast, we see that our approach of using a loss function based on topic distribution significantly improves the classifier performance. DistilBERT with its compact parameter size has the best F1 Score of $50.0\%$ for the granular labels task and $85.2\%$ for the coarse labels task.

\begin{table}[htbp]
\caption{Github-AI dataset granular class labels - F1 scores}
\label{tab_ghubaisub}
\centering
\begin{tabular}{lrrr}
\toprule
\textbf{Transformer} &  \textbf{No} & \textbf{LM} & \textbf{Our}\\
\textbf{} &  \textbf{Finetuning} & \textbf{Finetuning} & \textbf{Approach}\\
\midrule
BERT~\cite{devlin2018bert} & 22.8 & 27.2 & 45.0\\
DistilBERT~\cite{sanh2019distilbert} & 26.4 & 28.4 & 50.0\\
XLNet~\cite{yang2019xlnet} & 21.6 & 19.2 & 46.5\\
RoBERTa~\cite{liu2019roberta} & 21.3 & 19.9 & 34.4\\
Electra~\cite{clark2020electra} & 21.3 & 20.8 & 29.5\\
\bottomrule
\end{tabular}
\end{table}

\begin{table}[htbp]
\caption{Github-AI dataset coarse class labels - F1 scores}
\label{tab_ghubaisuper}
\centering
\begin{tabular}{lrrr}
\toprule
\textbf{Transformer} &  \textbf{No} & \textbf{LM} & \textbf{Our}\\
\textbf{} &  \textbf{Finetuning} & \textbf{Finetuning} & \textbf{Approach}\\
\midrule
BERT~\cite{devlin2018bert} & 68.9 & 66.7 & 80.8\\
DistilBERT~\cite{sanh2019distilbert} & 68.7 & 70.6 & 85.2\\
XLNet~\cite{yang2019xlnet} & 67.1 & 66.5 & 85.2\\
RoBERTa~\cite{liu2019roberta} & 67.2 & 66.7 & 69.9\\
Electra~\cite{clark2020electra} & 66.8 & 66.6 & 69.1\\
\bottomrule
\end{tabular}
\end{table}

\begin{table}[htbp]
\caption{Amazon dataset - F1 scores}
\label{tab_amazon1}
\centering
\begin{tabular}{lrrr}
\toprule
\textbf{Transformer} &  \textbf{No} & \textbf{LM} & \textbf{Our}\\
\textbf{} &  \textbf{Finetuning} & \textbf{Finetuning} & \textbf{Approach}\\
\midrule
BERT~\cite{devlin2018bert} & 32.0 & 78.1 & 86.5\\
DistilBERT~\cite{sanh2019distilbert} & 54.3 & 84.6 & 88.6\\
XLNet~\cite{yang2019xlnet} & 19.0 & 28.1 & 88.8\\
RoBERTa~\cite{liu2019roberta} & 20.8 & 73.3 & 88.2\\
Electra~\cite{clark2020electra} & 22.0 & 37.9 & 86.4\\
\bottomrule
\end{tabular}
\end{table}

\begin{table}[htbp]
\caption{Twitter dataset - F1 scores}
\label{tab_twitter1}
\centering
\begin{tabular}{lrrr}
\toprule
\textbf{Transformer} &  \textbf{No} & \textbf{LM} & \textbf{Our}\\
\textbf{} &  \textbf{Finetuning} & \textbf{Finetuning} & \textbf{Approach}\\
\midrule
BERT~\cite{devlin2018bert} & 22.2 & 40.1 & 46.0\\
DistilBERT~\cite{sanh2019distilbert} & 32.4 & 44.3 & 50.7\\
XLNet~\cite{yang2019xlnet} & 18.0 & 26.9 & 40.9\\
RoBERTa~\cite{liu2019roberta} & 15.0 & 29.3 & 22.7\\
Electra~\cite{clark2020electra} & 21.0 & 15.3 & 42.9\\
\bottomrule
\end{tabular}
\end{table}

\begin{figure}[!htbp]
\centering
\includegraphics[width=8cm]{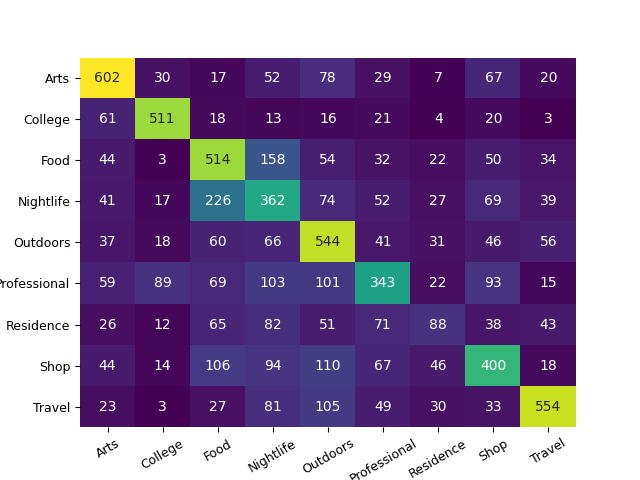}
\caption{Confusion Matrix for Twitter dataset.} 
\label{fig_cmtwitter}
\end{figure}

\begin{figure*}[!htbp]
\centering
\includegraphics[height=10cm]{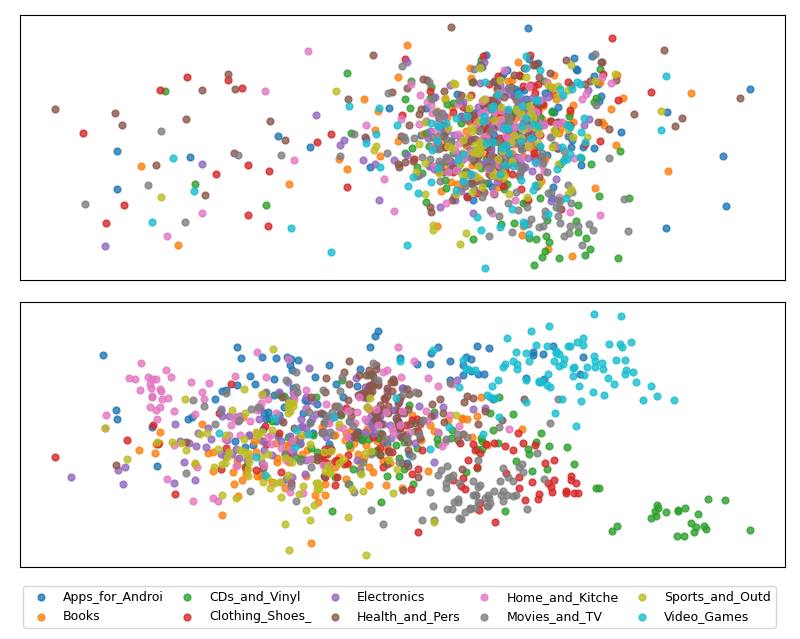}
\caption{Embedding Visualization in 2-D. \emph{top}: Embeddings produced using standard language model training objective. \emph{bottom}: Embeddings produced with loss function based on topic distribution. There is a perceptible grouping of points from the same class in the bottom picture.} 
\label{fig_embeds}
\end{figure*}

\begin{figure*}[!htbp]
\centering
\includegraphics[width=\textwidth]{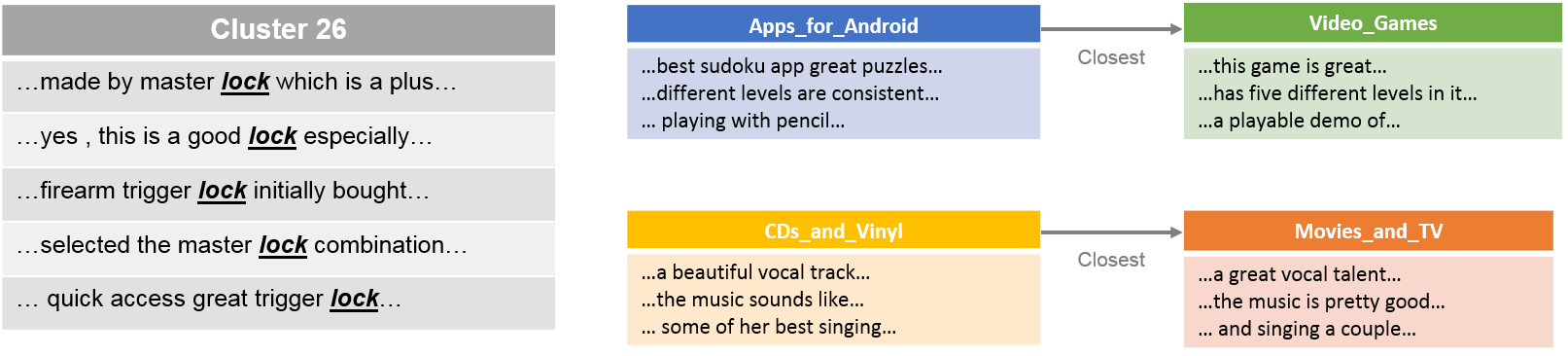}
\caption{Semantic composition in learned embeddings. \emph{left}: Discussions around \emph{lock} topic occur in the same cluster. \emph{right}: Semantically similar labels appear close together in the embedding space.}
\label{fig_clus}
\end{figure*}

\begin{figure}[!htbp]
\centering
\includegraphics[width=8cm]{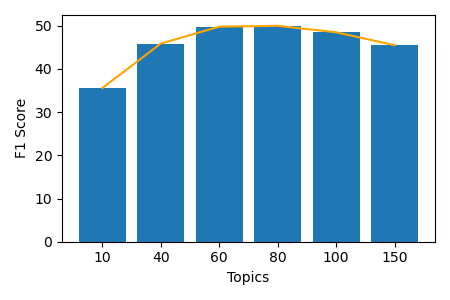}
\caption{Topic size effect on classifier performance.} 
\label{fig_topic}
\end{figure}

\begin{figure}[!htbp]
\centering
\includegraphics[width=8cm]{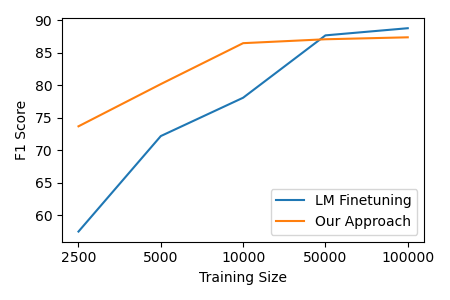}
\caption{Impact of training set size on classifier performance.} 
\label{fig_tsize}
\end{figure}

Table \ref{tab_amazon1} contains the results for Amazon dataset which has $10$ class labels. We see here that finetuning the language model significantly improves the classifier performance in general. This is unsurprising because unlike the GitHub-AI dataset, there are more examples to update the pre-trained weights. A noteworthy aspect is that our proposed training objective outperforms all the other baselines. The difference is less profound for models such as DistilBERT and BERT. The XLNet model performs the best with a F1 Score of $88.8\%$.

The results for Twitter dataset is presented in Table \ref{tab_twitter1} and the confusion matrix for the class labels is shown in Figure \ref{fig_cmtwitter}. The combination of informal language style in tweets, a small sample size and the large number of labels results in low F1 scores across all methods. Still, the DistilBERT model using our solution performs the best with a F1 score of $50.7\%$.

\subsection{Discussion}
It is important to understand why the proposed self-supervision based on topic modeling performs better over standard masked language modeling for the above classification tasks. Our approach aligns the input embeddings to reflect topic distributions, thereby partitioning the input space in a soft manner. This induces a clustering of input points with those points that share similar characteristics appearing together. The embeddings generated by standard language models do not necessarily have this clustering property. 

Figure \ref{fig_embeds} makes this partitioning effect evident. It plots in 2-D the inferred input embeddings for Amazon dataset. The top section of this figure contains the embeddings from a finetuned language model and there are no discernible clusters here. However, in the bottom section that corresponds to the same model trained using our objective function, we can clearly see patterns of points appearing together. This natural grouping of the inputs enable identification of class labels based on neighborhood search very effective.  

The learned embeddings also capture topic characteristics with input points corresponding to the same latent topic placed together. To validate this, we over-cluster the embeddings using KMeans and qualitatively examine the cluster contents. The left side of Figure \ref{fig_clus} contains one such cluster, where the discussions around \emph{locks} in the Amazon review are consolidated into the same cluster. Furthermore, semantically similar labels appear near to each other in the embedding space. We measure the distances between the cluster centroids and observe that the cluster closest to an \emph{Apps\_for\_Android} cluster is the \emph{Video\_Games} cluster. Similarly, a \emph{CDs\_and\_Vinyl} cluster is close to \emph{Movies\_and\_TV} cluster as illustrated in the right side of Figure \ref{fig_clus}. This level of compositionality opens the model up for applications to hierarchical classification and clustering.

For topic modeling using LDA~\cite{blei2003latent}, we need to specify the number of topics. This hyper-parameter influences the self-supervised loss function. We note that while it is essential to tune the number of topics, the model is not sensitive to an exact value. The effect of different topic size on the classifier performance for GitHub-AI dataset is plotted in Figure \ref{fig_topic}. Having extremely few topics does affect the model performance. However, the results are stable for a wide range of topic sizes.

Finally, we also study the impact of training set size on the proposed model. While the setting described here is intended for a sparse label scenario, we ask how the baseline model performs in the presence of large amounts of training data. Figure \ref{fig_tsize} compares the performance between our model and \emph{LM Finetuning} setup for the Amazon dataset. We see that our model performs significantly better when there is fewer training data available. However, F1 score for the finetuned model converges or gets better when training size increases beyond a threshold. We hypothesize that with large training set size, language semantics are understood better and the neighborhood search becomes easier.

\section{Conclusion}\label{sec_concl}
In this paper, we presented a flexible framework that combines latent topic information and metadata encodings with transformer-based models to learn semantically rich document representations that can be used for classification tasks in a transductive fashion. We demonstrate the flexibility of our framework by applying it to three datasets with diverse characteristics, various sizes, and different types of metadata. We show 4\%+ improvement over out-of-the-box pre-trained embeddings as well as conventional fine-tuning. We also qualitatively illustrate the semantic compositionality of the resulting embeddings. Our framework is especially effective when training data is smaller, when the classification task has a larger number of labels, or when metadata tags provide useful semantic signal that would otherwise be missed. In future work, we hope to explore the effectiveness of our framework in unsupervised hierarchical clustering. 

\begin{acks}
This paper was prepared for information purposes by the Artificial Intelligence Research group of JPMorgan Chase \& Co and its affiliates (``JP Morgan''), and is not a product of the Research Department of JP Morgan.  JP Morgan makes no representation and warranty whatsoever and disclaims all liability, for the completeness, accuracy or reliability of the information contained herein.  This document is not intended as investment research or investment advice, or a recommendation, offer or solicitation for the purchase or sale of any security, financial instrument, financial product or service, or to be used in any way for evaluating the merits of participating in any transaction, and shall not constitute a solicitation under any jurisdiction or to any person, if such solicitation under such jurisdiction or to such person would be unlawful. \copyright 2020 JPMorgan Chase \& Co. All rights reserved

\end{acks}

\bibliographystyle{ACM-Reference-Format}
\bibliography{main-acm}

\end{document}